\pdfoutput=1

\documentclass[11pt]{article}

\usepackage{acl}

\usepackage{times}
\usepackage{latexsym}
\usepackage{booktabs}
\usepackage{graphicx}
\usepackage{tabularx}
\usepackage{amsmath}
\usepackage{amssymb}

\usepackage[T1]{fontenc}

\usepackage[utf8]{inputenc}

\usepackage{microtype}

\title{Leveraging Non-dialogue Summaries for Dialogue Summarization}
\author{Seongmin Park\qquad Dongchan Shin\qquad Jihwa Lee\\
    ActionPower\\
  Seoul, Republic of Korea \\
  \texttt{\{seongmin.park, dongchan.shin, jihwa.lee\}@actionpower.kr} \\}

\begin{document}

\maketitle
\begin{abstract}
To mitigate the lack of diverse dialogue summarization datasets in academia, we present methods to utilize non-dialogue summarization data for enhancing dialogue summarization systems. We apply transformations to document summarization data pairs to create training data that better befit dialogue summarization. The suggested transformations also retain desirable properties of non-dialogue datasets, such as improved faithfulness to the source text. We conduct extensive experiments across both English and Korean to verify our approach. Although absolute gains in ROUGE naturally plateau as more dialogue summarization samples are introduced, utilizing non-dialogue data for training significantly improves summarization performance in zero- and few-shot settings and enhances faithfulness across all training regimes.
\end{abstract}

\section{Introduction}
Dialogue summarization fundamentally differs from its non-dialogue counterparts in two ways: the presence of speaker information and the inherent abstractiveness that demands any dialogue summarization system to "read between the lines". Consequently, training a dialogue summarization model requires datasets appropriate for the dialogue domain, which often calls for different provisions than those commonly found in traditional, non-dialogue summarization datasets. 

The bulk of research efforts in summarization, however, has historically been focused on written documents. As a result, the research community faces a shortage of diverse dialogue summarization data, in contrast to the abundance of non-dialogue summarization data \cite{feng2021survey, tuggener-etal-2021-summarizing}. From such state of the literature, we identify a strong need for methods to utilize widely available non-dialogue summarization data in reinforcing dialogue summarization models.

\begin{figure}[th]
  \centering
  \includegraphics[width=\linewidth]{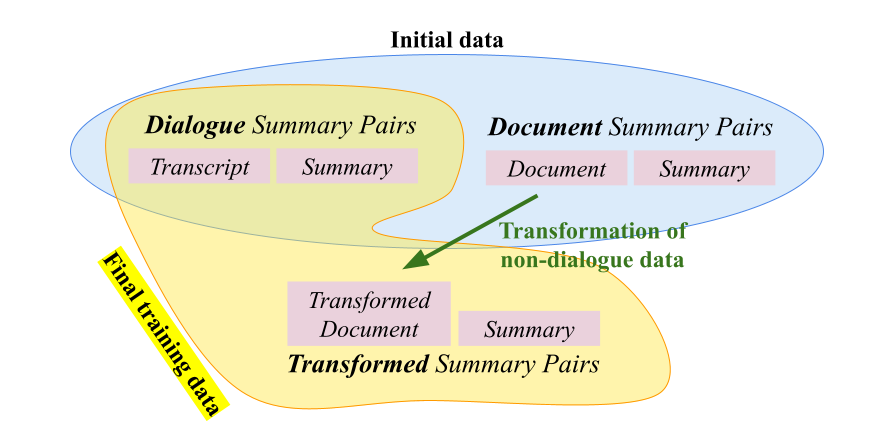}
  \caption{Overview of our proposed method. We transform non-dialogue data into a format exploitable by dialogue summarization models.}
  \label{fig:overview}
\end{figure}
In this work, we present recipes to transform non-dialogue data into formats that enable direct integration into dialogue summarization training. During the transformation process, we also inherit desirable properties that arise from the extractiveness of non-dialogue summarization datasets. Factual inconsistency and hallucination are major research problems in dialogue summarization \cite{maynez-etal-2020-faithfulness, ladhak2021faithful, cao2018faithful, huang2021factual}. Since extractive summaries naturally remain more faithful to the source text, we design our transformation schemes to retain such properties when adapting non-dialogue summary data to the dialogue domain. 

Our contributions are as follows:
\begin{enumerate}
  \item We present formulas to transform non-dialogue summarization datasets into patterns usable for dialogue summarization. Summarization models trained with the additional data produce summaries more similar to gold reference summaries.
  \item We show that utilizing non-dialogue summarization data preserves faithfulness in otherwise factually-unchecked summaries. 
  \item We test our data manipulation scheme across two languages (English and Korean) and on document summary datasets with different levels of abstraction.
\end{enumerate}

In Section \ref{sec:background}, we first describe existing challenges in dialogue summarization. In Section \ref{sec:methods}, we describe our dataset adaptation methods in detail. In Section \ref{sec:experiments}, we describe datasets, evaluation metrics, and experiments used to test our methods.

\section{Related works} \label{sec:background}
\subsection{Non-dialogue data for dialogue summarization}
Even in high-resource languages like English, diversely annotated dialogue summarization datasets are scarce \cite{feng2021survey, tuggener-etal-2021-summarizing, zou-etal-2021-low}. The need for a diverse collection of dialogue summarization datasets is further exacerbated by the fact that dialogue is recorded in many formats, such as meetings, chats, and spontaneous speech. 

To appease such a need for more data, several attempts have been made to utilize non-dialogue data in dialogue summarization (Figure \ref{fig:overview}). \cite{zou-etal-2021-low} pre-trains a language model with BookCorpus \cite{7410368} to provide training samples across diverse domains. \cite{khalifa-etal-2021-bag} pre-trains BART \cite{lewis-etal-2020-bart} with unlabeled dialogue corpora and fine-tunes the language model with downstream summary tasks. 

The focus of such approaches lies in whetting a model to be more responsive to limited dialogue summarization data. We suggest a new line of research that directly manipulates the training data instead of steering a model's disposition directly.

\subsection{Faithfulness in dialogue summarization}
Factual incorrectness is a problem commonly observed in abstractive summarization systems \cite{cao2018faithful, huang2021factual, tang2021confit}. \citet{tang2021confit} identifies categories of factual errors that dialogue summarization models may generate. To improve the factual consistencies of generated summaries, the authors corrupt dialogue transcripts to create negative samples in a contrastive-learning scheme. 

We employ a similar noising approach. Negative sample generation in \cite{tang2021confit} requires accurate token-level operations, such as part-of-speech extraction and word negation. Our manipulation scheme forgoes such additional components, relying only on deterministic sentence-level edits.

\begin{figure}[t]
  \centering
  \includegraphics[width=\linewidth]{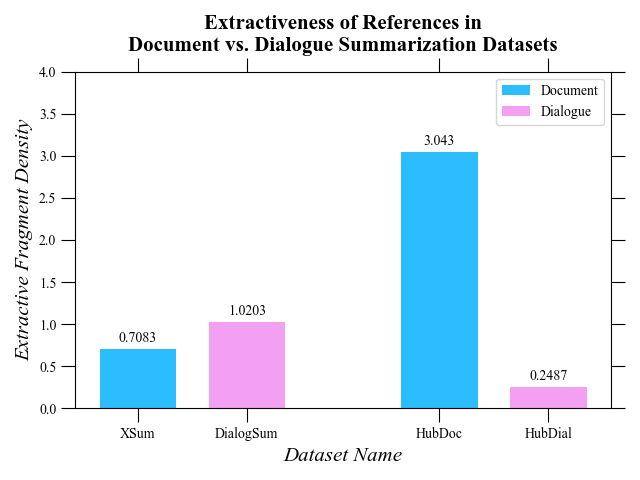}
  \caption{Extractiveness of reference summaries. Extractive Fragment Density \cite{grusky-etal-2018-newsroom} is the longest extractive token span from the source data that matches the reference summary.}
  \label{fig:extractiveness}
\end{figure}
\begin{table*}[th]
\small
  \caption{Evaluation metrics for full training. $f_d$ (D) transformation is consistently effective in boosting match-based ROUGE. Even though marginal gain in ROUGE from our method naturally decreases as the size of $DialSet$ increases, incorporating document summarization data greatly improves summary faithfulness. R1, R2, RL, Prec., Rec., Faith. respectively stands for ROUGE-1, ROUGE-2, ROUGE-L, Precision, Recall, and Faithfulness. Underlined values are the highest in each column. Higher is better for all metrics.}
  \label{tab:full}
  \centering
  \begin{tabular}{l||c|c|c|c|c|c||c|c|c|c|c|c}
    \toprule
    \textbf{Data} & \multicolumn{6}{c}{DialogSum} & \multicolumn{6}{c}{HubDial}   \\   
    \midrule
     & \textit{R1} & \textit{R2} & \textit{RL} & \textit{Prec.} & \textit{Rec.} & \textit{Faith.} & \textit{R1} & \textit{R2} & \textit{RL} & \textit{Prec.} & \textit{Rec.} & \textit{Faith.}                 \\
     \midrule
    Original & 39.57 &15.43 &	32.97 & -3.8962 & -4.3175 & -4.1101 & 35.42 & 16.90 & 31.11 &  -9.9774 & -9.7934 & -8.9965\\
     Naive& 39.36 & 14.89 & 32.56 & -2.9108 & -2.9933 & -2.5351 & 35.97 & 17.68	& 31.13 & -7.8063 & -7.6203 & -7.9189\\
     D & \textbf{40.47} & \textbf{16.41} & \textbf{33.89} & \textbf{-2.9085} & -2.8702	& -2.4615 & \underline{\textbf{36.32}} & 17.61 & \textbf{31.79} & \underline{\textbf{-7.7715}} & -7.6190 & -7.9202\\
     S & 39.94 &	15.70 &	33.31 & -2.9310 & -2.8966 & -2.4261 & 36.08 & \underline{\textbf{18.13}}	& 31.41 & -7.8159 & \underline{\textbf{-7.5791}} & \underline{\textbf{-7.8610}}\\
     O & 39.80 & 15.97 & 33.32 & -2.9211 & \textbf{-2.8253} & \underline{\textbf{-2.4072}} & 35.96 & 17.51	& 31.32 & -7.7975 & -7.6251 & -7.8954\\
     \midrule
     D + S & 39.87 &	15.73 & 33.44 & -2.9235 & -2.9224 &  -2.4970 & 36.03 & \textbf{17.55}	& \underline{\textbf{31.93}} & -7.8179 & -7.6066 & \textbf{-7.9104}\\
     D + O  & \underline{\textbf{40.66}} &	\underline{\textbf{16.77}} & \underline{\textbf{34.15}} & \underline{\textbf{-2.9044}} & \underline{\textbf{-2.8073}} & \textbf{-2.4196} & 36.11 & 17.52 & 31.92 & \textbf{-7.7776} & -7.6245 & -7.9395\\
     S + O & 40.34 & 16.33 & 33.82 & -2.9077 & -2.8797 & -2.4376 & 35.52 & 17.21	& 31.32 & -7.8456 & -7.6149 & -7.9244\\
     D + S + O & 39.97 &	16.00 & 33.56 & -2.9402 & -2.8678 & -2.4278 & \textbf{36.26} & 17.29 & 31.29 & -7.8105 & \textbf{-7.5956} & -7.9371\\
    \bottomrule
  \end{tabular}
\end{table*}

\section{Proposed method} \label{sec:methods}

\subsection{Preliminaries}
Let \begin{equation}
DocSet = \{(A_0, X_0), (A_1, X_1), ... , (A_i,X_i)\}
\end{equation} 
be a non-dialogue (document) summarization dataset, where $A_i$ is the $i$-th document in the set, and $X_i$ is the corresponding reference summary. $A_i$ is a sequence of sentences $(a_0, a_1, ..., a_m)$, where $m$ is the sentence count.

Similarly, we define a dialogue summarization dataset, 
\begin{equation}DialSet = \{(B_0, Y_0), (B_1, Y_1), ... , (B_j,Y_j)\},\end{equation} where $B_j$ is the $j$-th dialogue transcript in the dataset, and $Y_j$ is the corresponding dialogue summary. Like any $A$, $B_j$ consists of ordered sentences $(b_0, b_1, ..., b_n)$.

We define $F = \{{f_0, ..., f_k}\}$, a set of \textit{transformation functions} to be applied to each $A_i$ in $DocSet$. A transformation function is a set of operations to transform non-dialogue text data into a pattern usable in dialogue summarization training. 

We introduce three such transformation functions: \textbf{forcing plain text into dialogue format} (e.g. by inserting pseudo-speaker information), \textbf{shuffling sentence order}, and \textbf{omitting the sentence with highest extractive overlap} with the reference summary. Each suggested transformation function is formerly defined in succeeding sections.

Once $f_k$ is applied to each $A_i$ in $DocSet$, each transformed non-dialogue input text is paired with its corresponding reference $X_i$ to form a new training set:
\begin{equation}
NewDocSet = \{(f_k(A), X) \mid (A, X) \in DocSet \}.
\end{equation}
$NewDocSet$ can be used as additional training data for dialogue summarization models.

\subsection{Arranging text into dialogue format ($f_d$)}
Given a plain document, we convert its contents into transcript format by segmenting the document into sentences and appending a psuedo speaker:
\begin{equation}
f_d(A) = (concatenate(``Speaker\ \textit{1}: ",\ a))_{a \in A}.\end{equation}
This operation serves two purposes: we prime our model to be more receptive of dialogue-formatted data through prompting \cite{liu2021pre}. We also remove the gap between data patterns in training and inference by standardizing diverse non-dialogue document data into the dialogue domain.   

The prompt \textit{"Speaker 1"} was chosen empirically: multiple configurations, such as varying speaker numbers and inserting real names, were tested. Such complex configurations led to only marginal increases in evaluation metrics and introduced additional roadblocks in reliable reproduction (upper bound in speaker number has to be arbitrarily selected; a dictionary with realistic names has to be distributed). Both English and Korean datasets used \textit{"Speaker 1"}.

\subsection{Shuffling sentence order ($f_d$)}
To combat lead bias commonly observed in traditional summarization datasets \cite{grenander-etal-2019-countering, zhu2021leveraging}, we shuffle the order of sentences in A:
\begin{equation}
f_s(A) = shuffle(A).
\end{equation}
 Previous research has shown sentence shuffling helps in reducing read bias \cite{grenander-etal-2019-countering}. Since information in dialogues is often dispersed across multiple utterances, we find sentence shuffling to be more impactful when dealing with dialogues, compared to documents.  

\subsection{Omitting the most extractive sentence ($f_o$)}
Among all sentences in a document, we delete the sentence with the most extractive overlap with the reference summary. The degree of overlap is calculated by the number of shared character 3-grams between a single sentence from the source document and the whole reference.
\begin{equation}
f_o(A_i) = A_i \smallsetminus \{a_{ex}\}, 
\end{equation}
where $a_{ex}$ in $A_i$ has the highest 3-gram overlap with $X_i$.
By removing the most extractive sentence, we aim to make $DocSet$ more abstractive and reduce copying behavior.

\begin{table}[h]
  \small
  \caption{Datasets used in the experiment. ``Dial." and ``Doc." stand for ``dialogue" and ``document".}
  \label{tab:datasets}
  \centering
  \begin{tabular}{c|c|c|c|c}
    \toprule
    \textbf{Name} & \textbf{Lang.} &\textbf{Type} &\textbf{Size}&\textbf{Abstractive?}\\   
    \midrule
    DialogSum & English & Dial. & 15,600 & Yes \\
    XSum & English & Doc. & 204,045 & Yes \\
    HubDial & Korean & Dial. & 16,000 & Yes \\
    HubDoc & Korean & Doc. & 334,160 & No \\
    \bottomrule
  \end{tabular}
\end{table}

\section{Experiments} \label{sec:experiments}

\subsection{Experiment setup}
We conduct comprehensive experiments that apply transformation functions defined in Section \ref{sec:methods}. \subsubsection{Our models}
First, we create different variants of $NewDocSet$ by applying functions in $F = \{f_d, f_s, f_o\}$ both individually and in combination. Such application results in 7 different variations of $NewDocSet$: $D$, $O$, $S$, $D+O$, $D+S$, $S+O$, $D+S+O$, where, for example, $D+O = \{f_d \circ f_o(A) \mid A \in DocSet\}$.

With newly acquired training data, we train a BART-base \cite{lewis-etal-2020-bart, wolf2019huggingface} summarizer under three different configurations:
\begin{enumerate}
  \item \textit{Zero-shot}: $NewDocSet$ is the training set.
  \item \textit{Few-shot}: Training data consists of $NewDocSet$ and 100 or 1000 samples from $DialSet$.
  \item \textit{Full training}: Training data consists of $NewDocSet + DialSet$.
\end{enumerate}
We choose the BART architecture due to its widespread use and proven track record in summarization \cite{fabbri2021convosumm, akiyama-etal-2021-hie, zhao2021todsum}.
\subsubsection{Baselines}
We compare our trained models with two baselines: 
\begin{enumerate}
 \item \textit{Original}: $DialSet$ is the training set.
  \item \textit{Naive}: Training data consists of $DialSet$ and $DocSet$ (i.e. $f_{naive}(A) = A$).
\end{enumerate}

\subsection{Datasets}

For English, we use DialogSum \cite{chen-etal-2021-dialogsum} as $DialSet$ and XSum \cite{xsum-emnlp} as $DocSet$. For Korean, we use AIHub Dialogue Summarization Dataset\footnote{https://aihub.or.kr/aidata/30714} (HubDial) as $DialSet$ and AIHub Document Summarization Dataset\footnote{https://aihub.or.kr/aidata/8054} (HubDoc) as $DocSet$. Table \ref{tab:datasets} contains a brief description of each dataset.  

Transformations $f_s$ and $f_o$ hinge on the assumption that non-dialogue summarization datasets typically display considerable lead bias and are more extractive than dialogue summarization datasets. To gauge how extractiveness of non-dialogue data effects final summary generation performance, we conduct experiments on both highly extractive (HubDoc) and extremely abstractive (XSum) document summarization datasets (Figure \ref{fig:extractiveness}).

\subsection{Evaluation metrics}

Performance of our model is measured as the similarity between model summaries and reference summaries, calculated with standard ROUGE scores (ROUGE-1, ROUGE-2, and ROUGE-L) \cite{lin-2004-rouge}. We also measure the faithfulness of the output summaries to input dialogues with BartScore \cite{yuan2021bartscore}. BartScore is a state-of-the-art evaluation metric for factual consistency and faithfulness in text generation. 

\begin{table*}[t]
  \small
  \caption{Few-shot results on English DialogSum. Since XSum is already highly abstractive, $f_d$ (D) transformation is the most effective. Almost all maximum values in each category involve a $f_d$ transformation. Notations are the same as in Table \ref{tab:full}.}
  \label{tab:english}
  \centering
  \begin{tabular}{l||c|c|c|c||c|c|c|c||c|c|c|c}
    \toprule

    & \multicolumn{4}{c}{Zero-shot} & \multicolumn{4}{||c||}{100-shot}  & \multicolumn{4}{c}{1000-shot} \\ 
    \midrule
     & \textit{R1} & \textit{R2} & \textit{RL} & \textit{Faith.} & \textit{R1} & \textit{R2} & \textit{RL} & \textit{Faith.} & \textit{R1} & \textit{R2} & \textit{RL} & \textit{Faith.}                 \\
     \midrule
    Original& -	&-&	- &-& 31.05&	10.55&	26.58&	-4.4925& 35.12	&12.23&	29.20& -4.7314\\
     Naive &13.64&	2.71&	11.21 &-2.9081 & 31.05	&9.31	&25.81&-4.5829 &37.97	&13.22&	31.11& -2.4799\\
     D &15.46	&3.18	&13.05& \textbf{-2.7045} & \underline{\textbf{34.93}}&	\underline{\textbf{10.74}}&	\underline{\textbf{28.21}}&	-4.4414
&38.28&	13.23&	31.08& -2.4319\\
     S &14.35&	\textbf{3.51}&	12.11& -2.8926 & 32.83&	09.82&	26.80	&\underline{\textbf{-2.4675}}&\textbf{38.33}	&13.62	&\textbf{31.45} &\textbf{-2.4146} \\
     O &\textbf{16.41}&	2.80&	\textbf{13.66} &-2.9846 & 32.89&	09.53&	27.24&	-2.8102&38.28	&\underline{\textbf{13.57}}&	31.12 &-2.4495\\
     \midrule
     D + S &\underline{\textbf{17.47}}&	\underline{\textbf{4.27}}&	\underline{\textbf{14.56}} &\underline{\textbf{-2.4899}} & 34.51&	\textbf{10.96}	&27.96	&-2.7527 & 38.26&	13.27	&31.21& -2.4494\\
     D + O  &14.73	&2.88&	12.07 &-2.9530& 34.40	&10.76	&28.18	&\textbf{-2.6696}&\underline{\textbf{38.85}}	&\textbf{13.55}&	\underline{\textbf{31.62}}& \underline{\textbf{-2.3949}}\\
     S + O &16.69&	3.42&	13.96 &-3.1872 & 33.84& 10.27&	27.83&-3.0668 & 38.55&	13.44	&31.19& -2.4154
\\
     D + S + O &16.36	&3.84	&13.76& -2.5456 & \textbf{34.65}&	10.82&	\textbf{28.25}&	-2.7152& 36.80&	13.03	&30.42& -2.4381\\
    \bottomrule
  \end{tabular}
\end{table*}

\begin{table*}[t]
  \small
  \caption{Few-shot results on Korean HubDial. Compared to less extractive English summarization, we see $f_s$ (S) and $f_o$ (O) transformations resulting in greater marginal increase in ROUGE. Notations are the same as in Table \ref{tab:full}.}
  \label{tab:korean}
  \centering
  \begin{tabular}{l||c|c|c|c||c|c|c|c||c|c|c|c}
    \toprule

    & \multicolumn{4}{c}{Zero-shot} & \multicolumn{4}{||c||}{100-shot}  & \multicolumn{4}{c}{1000-shot} \\ 
    \midrule
     & \textit{R1} & \textit{R2} & \textit{RL} & \textit{Faith.} & \textit{R1} & \textit{R2} & \textit{RL} & \textit{Faith.} & \textit{R1} & \textit{R2} & \textit{RL} & \textit{Faith.}                 \\
     \midrule
    Original& -	&-&	- &-& 3.41&	1.35&	3.03&	-10.2144& 31.42&	13.64	&26.69& -7.9586\\
     Naive &20.72& 8.94&	18.34  &-7.4637  & 27.98&	12.56&	24.24&-7.7524 &32.17&	14.74&	27.34& -7.8893\\
     D &\underline{\textbf{26.34}}&\underline{\textbf{11.74}}&\underline{\textbf{22.97}}& -7.6731& 28.48&	13.03&	24.79&	\underline{\textbf{-7.6451}}&\underline{\textbf{33.05}}&	\textbf{15.16}	&\underline{\textbf{28.46}}& \underline{\textbf{-7.7255}}\\
     S &21.38&	9.43&	19.01& -7.3379 & 28.12&	12.42	&24.19	&-7.9053&32.68	&14.91	&27.78 &-7.8162 \\
     O &22.09&	9.82&	19.73 &\underline{\textbf{-7.1363}} & \textbf{29.50}	&\textbf{13.26}	&\textbf{25.01}&	-7.8737&32.26&	14.77&	27.85 &-7.8357\\
     \midrule
     D + S &24.21&	11.31&	21.48 &-7.8747 & 28.50	&13.06	&24.84	&-7.8565 & 31.78&	14.62&	27.18& -7.7687\\
     D + O  &\textbf{24.81}	&11.20&	\textbf{22.04} &-7.7556 & 29.71&	13.17&	\underline{\textbf{25.68}}	&\textbf{-7.8058} &31.67&	14.66	&27.33& \textbf{-7.7651}\\
     S + O &20.38&	9.25	&18.50 &\textbf{-7.4731} & \underline{\textbf{29.79}}&	\underline{\textbf{13.47}}&	25.62&-7.9142 & 31.92&	14.53&	27.16& -7.7673\\
     D + S + O &24.17&	\textbf{11.37}&	21.46& -7.8234 & 28.50	&13.31	&24.48&	-7.8908& \textbf{32.77}	&\underline{\textbf{15.25}}&	\textbf{27.96}& -7.7934\\
    \bottomrule
  \end{tabular}
\end{table*}

\begin{figure}[h]
  \centering
  \includegraphics[width=\linewidth]{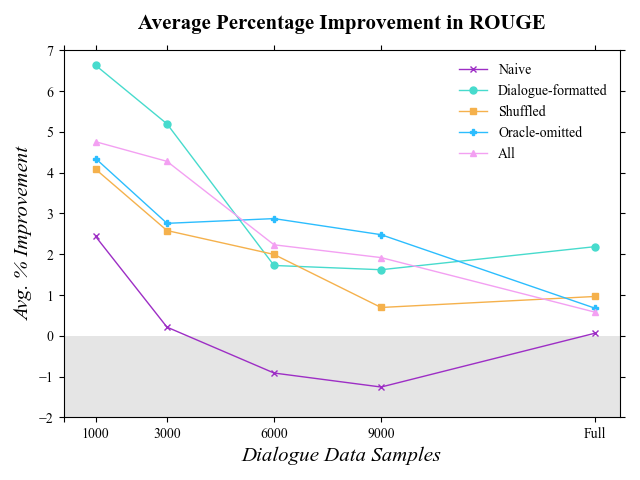}
  \caption{Averaged percentage ROUGE-1, ROUGE-2, and ROUGE-L improvements over dialogue-only training on HubDial test set. Shaded regions indicate configurations that underperform dialogue-only training.}
  \label{fig:korean_few}
\end{figure}

\section{Results}

\subsection{Full training}
Both English and Korean summarization models benefit from additional data curated by our transformation functions. Only naive application of non-dialogue data fails to improve ROUGE scores compared to dialogue-only training. While marginal increase in ROUGE saturates as more dialogue summaization training samples are added, the addition of document data significantly enhances factual consistency of summaries (Table \ref{tab:full}). 
\subsubsection{Abstractive document summary dataset}
In terms of ROUGE, models trained with abstractive document summarization data (XSum) are most affected by $f_d$ (D) transformations. Highest scoring data transformation combinations mostly involve $f_d$. 
In terms of factual consistency and faithfulness, $f_o$ transformations consistently score the highest. This is in line with our intention to introduce an additional in-comprehension understanding objective to the model that simple dialogue formatting cannot provide. 
\subsubsection{Extractive document summary dataset}
$f_s$ (S) and $f_o$ (O) transformations are more influential when used to transform extractive data (HubDoc). Factual consistency is correlated the most with $f_s$, because of lead bias present in HubDoc.

\subsection{Zero- and few-shot training}
In zero- and full-shot training, we see significant improvements in both ROUGE and factual consistency (Tables \ref{tab:english}, \ref{tab:korean}). Figure \ref{fig:korean_few} shows improvements in ROUGE over $DialSet$-only training at different dialogue $DialSet$ sizes. Naively training with non-dialogue summarization data yields results no better than training with only dialogue data. In contrast, our suggested transformations provide significant gains in both span match and consistency measures in low-shot training regimes. 

Comparative influence of each transformation function ($f_d$, $f_s$, and $f_o$) show trends similar to those observed in full training, with $f_d$ proving the most dominant for already abstractive $DocSet$ (XSum) and $f_s$ and $f_o$ being more influential in comparatively extractive $DocSet$ (HubDoc).

\section{Conclusion}

We present simple but immediately effective methods to utilize abundant non-dialogue summarization data to improve dialogue summarization systems. We evaluate performance gains in similarity to reference summaries as well as in factual consistency to original transcript input. We find that our method is especially impactful in low-resource dialogue summarization.

Our research hints at two possible avenues for further investigation: reinforcing the three presented transformation recipes with a more methodical generation of prompts \cite{DBLP:journals/corr/abs-2112-05717}, or introducing new transformations that better capture the unique properties of dialogue summarization datasets.

\bibliographystyle{acl_natbib}

\bibliography{mybib}


\end{document}